\documentclass[10pt,twocolumn,letterpaper]{article}

\usepackage[pagenumbers]{cvpr} 

\usepackage{multirow}
\usepackage{algorithm}
\usepackage{algpseudocode}
\definecolor{cvprblue}{rgb}{0.21,0.49,0.74}
\usepackage[pagebackref,breaklinks,colorlinks,allcolors=cvprblue]{hyperref}

\def\confName{CVPR}
\def\confYear{2027}

\title{OmniHarness-RSI: Modality-Aware Harness Evolution for Omnimodal Understanding}

\author{Shilin Ma\thanks{Equal contribution.}, Xingming Shui\footnotemark[1], Jinglin Chen, Zetong Zhao\\
Jingqi Tian, Deyi Zhu, Yansong Tang\\
Tsinghua Shenzhen International Graduate School, Tsinghua University
}

\begin{document}
\maketitle
\begin{abstract}
    Recent advances in harness-based recursive self-improvement (RSI) have improved video understanding by evolving executable agent workflows without updating foundation model parameters. However, extending this paradigm to omnimodal understanding remains challenging, as the heterogeneous and interdependent nature of multimodal perception and reasoning creates a complex evolution space, making targeted harness optimization difficult. To address these challenges, we propose \textbf{OmniHarness-RSI}, a modality-aware framework for evolving agent harnesses. Specifically, we factorize the evolution space into three capability dimensions: modality routing and collaboration, modality-aware evidence localization, and modality-specific perception. Based on this factorization, we construct a Factorized-Harness Agent integrating specialized perception experts and audio-video retrieval tools. We further introduce modality-guided harness evolution, which identifies deficiencies through structured modality-capability attribution and refines corresponding harness strategies through interactive code revisions and execution feedback. Validation-guided selection and held-out evaluation assess generalization within and across datasets. Experiments on WorldSense, Daily-Omni, and OmniVideoBench demonstrate test accuracy improvements of 8.86, 8.13, and 6.23 percentage points, respectively, consistently outperforming modality-agnostic RSI. The evolved harnesses also demonstrate cross-dataset transfer without adaptation, while reducing planner calls and viewed frames.
    \end{abstract}    
\section{Introduction}
\label{sec:intro}

\begin{figure*}[t]
    \centering
    \includegraphics[width=\textwidth]{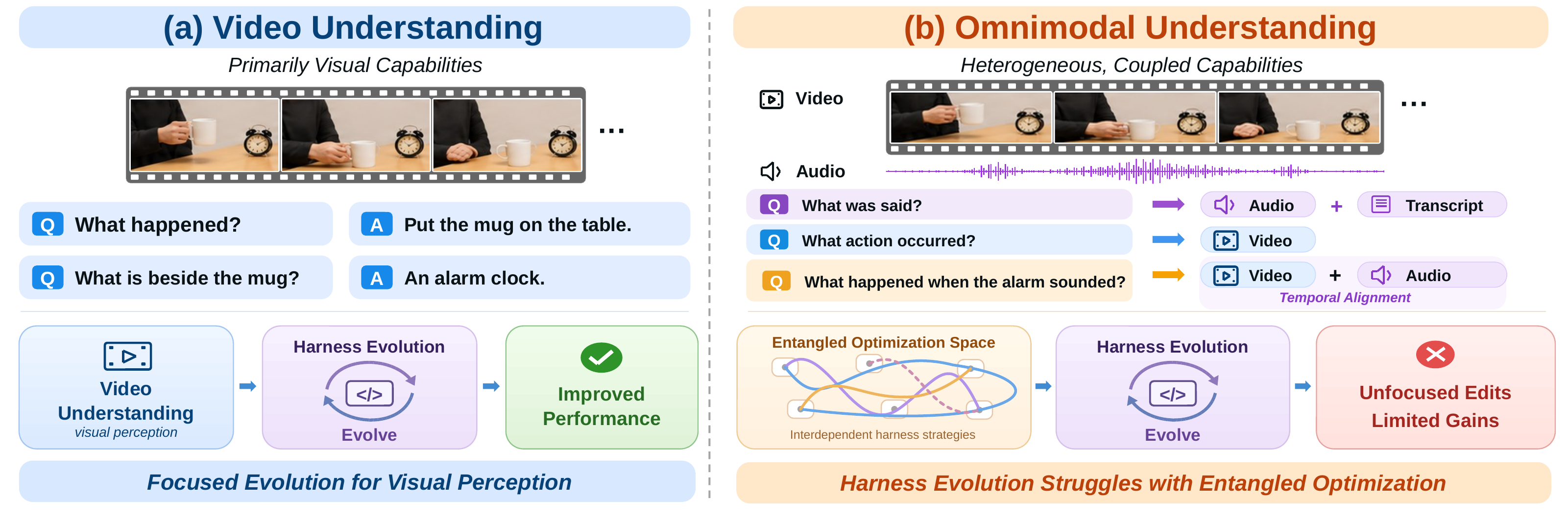}
    \caption{\textbf{Harness evolution for video and omnimodal understanding.}
    (a) Video-centric tasks primarily emphasize visual perception, providing a relatively focused space for harness evolution.
    (b) Omnimodal questions require different combinations of audio, visual, and textual evidence, often with temporal alignment.
    The resulting interactions among harness strategies make failure attribution and targeted refinement more challenging, motivating modality-aware harness evolution.}
    \label{fig:comparison}
\end{figure*}

Harness-based recursive self-improvement (RSI) improves agent performance by optimizing executable perception and reasoning workflows while keeping foundation model parameters frozen~\cite{lee2026rhi,lin2026ahe,wu2026modularrsi,xia2026rrsi}. Recent methods refine context construction, evidence acquisition, and agent workflows through code revisions and execution feedback~\cite{xu2026videoharnessrsi,luo2026videorsi,xu2026videoevoharness} in video understanding. These studies highlight the potential of harness evolution to improve agent capabilities by refining how frozen models and external tools are orchestrated for evidence acquisition and reasoning.

Despite these advances, extending harness-based RSI to omnimodal understanding remains challenging. As illustrated in Fig.~\ref{fig:comparison}, omnimodal questions require different combinations of audio, visual, and textual evidence, calling for task-dependent modality routing and cross-modal collaboration~\cite{xu2025qwen3omni,tao2025omniagent}. For instance, speech-related questions rely heavily on audio perception, visual event recognition requires detailed video analysis, and cross-modal reasoning may require temporally aligned evidence from both. Harness evolution must therefore account for coupled strategies governing expert routing and collaboration, evidence localization, and modality-specific perception. Their interactions create an entangled evolution space, making failure attribution ambiguous and the effects of unfocused modifications difficult to assess.

To address these challenges, we propose \textbf{OmniHarness-RSI}, a modality-aware harness evolution framework for omnimodal understanding. Our key insight is \textbf{modality-aware evolution space factorization}, which organizes editable harness strategies into three complementary capability dimensions: \textbf{modality routing and collaboration}, \textbf{modality-aware evidence localization}, and \textbf{modality-specific perception}. These dimensions govern which modalities and experts to invoke and how to coordinate them, where to locate question-relevant evidence, and how to interpret that evidence, respectively. Based on this factorization, we construct a \textbf{Factorized-Harness Agent} that integrates specialized perception experts and audio-video retrieval tools within a shared executable harness. This structure links the capability dimensions to concrete editable components, enabling targeted and coordinated revisions. Building on this factorization, we introduce \textbf{modality-guided harness evolution}, comprising \textbf{structured modality-capability attribution} and \textbf{capability-aligned harness refinement}. Rather than treating failures as general workflow deficiencies, our evolution agent identifies the affected modalities and specific harness strategies, enabling fine-grained diagnosis of modality-specific perception and cross-modal coordination failures. Guided by these attributions, the agent performs targeted code revisions to address recurring failure patterns, using interactive local trials to assess corrected errors and potential regressions. This process connects modality-aware failure diagnosis with capability-aligned optimization, enabling more focused and effective evolution. Furthermore, we introduce a \textbf{generalization-aware evolution and evaluation protocol} with disjoint evolution, validation, and test sets. Harness candidates are selected based on validation performance, while generalization is assessed at three levels: evolution-set improvement, in-distribution generalization, and cross-dataset transfer without further adaptation. This protocol distinguishes evolution-set improvement, in-distribution generalization, and cross-dataset transfer.

Experiments on WorldSense, Daily-Omni, and OmniVideoBench show that our modality-guided RSI improves held-out test accuracy by 8.86, 8.13, and 6.23 percentage points over the initial Factorized-Harness Agent, respectively, and consistently outperforms modality-agnostic RSI on the same architecture. The WorldSense-evolved harness transfers to Daily-Omni and OmniVideoBench without further evolution, yielding test gains of 7.67 and 12.10 points over the initial agent, respectively, with performance comparable to or better than harnesses evolved directly on the target datasets. Inference analysis further shows consistent reductions in planner calls and viewed frames across all three benchmarks, indicating more efficient use of planning interactions and visual inputs.

\section{Related Work}
\label{sec:related_work}

\subsection{Omnimodal Understanding}

Recent advances in omnimodal large language models have enabled unified perception and reasoning across audio, visual, and textual modalities~\cite{cheng2024videollama2,sun2024videosalmonn,fu2024vita,li2025baichuanomni15,xu2025qwen25omni,xu2025qwen3omni}. Beyond unified architectures, recent studies further enhance audio-visual reasoning through reinforcement learning, cross-modal representation alignment, and modality-aware training strategies~\cite{sun2025videosalmonno1,xing2025echoinkr1,chen2026omnivideor1,jia2026omnijigsaw,liu2026omniboost,xu2026syncra}. Meanwhile, emerging benchmarks systematically evaluate fine-grained audio-visual understanding, temporal alignment, and cross-modal reasoning~\cite{li2025omnibench,zhou2025dailyomni,hong2026worldsense,li2026omnivideobench,tao2026lvomnibench,wang2026avibench,chao2026jointavbench}. Complementing model-centric approaches, agent-based methods leverage adaptive perception and tool orchestration to acquire and integrate multimodal evidence~\cite{tao2025omniagent,chen2026omnireasoner,wu2026sleuthvlm}. Despite these advances, the automated optimization of modality-specific perception and cross-modal collaboration through executable harness evolution remains relatively underexplored. Our work addresses this direction through modality-aware evolution space factorization and targeted harness refinement without updating foundation model parameters.

\subsection{Harness Evolution for Video Understanding}

Recent studies have explored harness-based recursive self-improvement (RSI) as a promising paradigm for enhancing video understanding without modifying foundation model parameters. VideoHarness-RSI~\cite{xu2026videoharnessrsi} introduces automated evolution of executable context-construction pipelines, while Video-RSI~\cite{luo2026videorsi} leverages active video investigation and execution feedback to refine evidence acquisition strategies. VideoEvoHarness~\cite{xu2026videoevoharness} jointly evolves question generation and solving harnesses from unlabeled videos. More recently, VideoHarness-SRSI~\cite{lin2026videoharnesssrsi} investigates overfitting and unreliable feedback in recursive harness optimization, introducing structured search mechanisms to improve generalization. Despite these advances, existing methods primarily investigate harness evolution in video-centric settings, while the heterogeneous expert capabilities and complex cross-modal dependencies in omnimodal understanding introduce additional optimization challenges. Our OmniHarness-RSI addresses these challenges through capability-structured evolution, targeted interactive refinement, and generalization-guided harness selection.

\section{Method}
\label{sec:method}

\begin{figure*}[t]
    \centering
    \includegraphics[width=\textwidth,trim=10bp 60bp 10bp 80bp,clip]{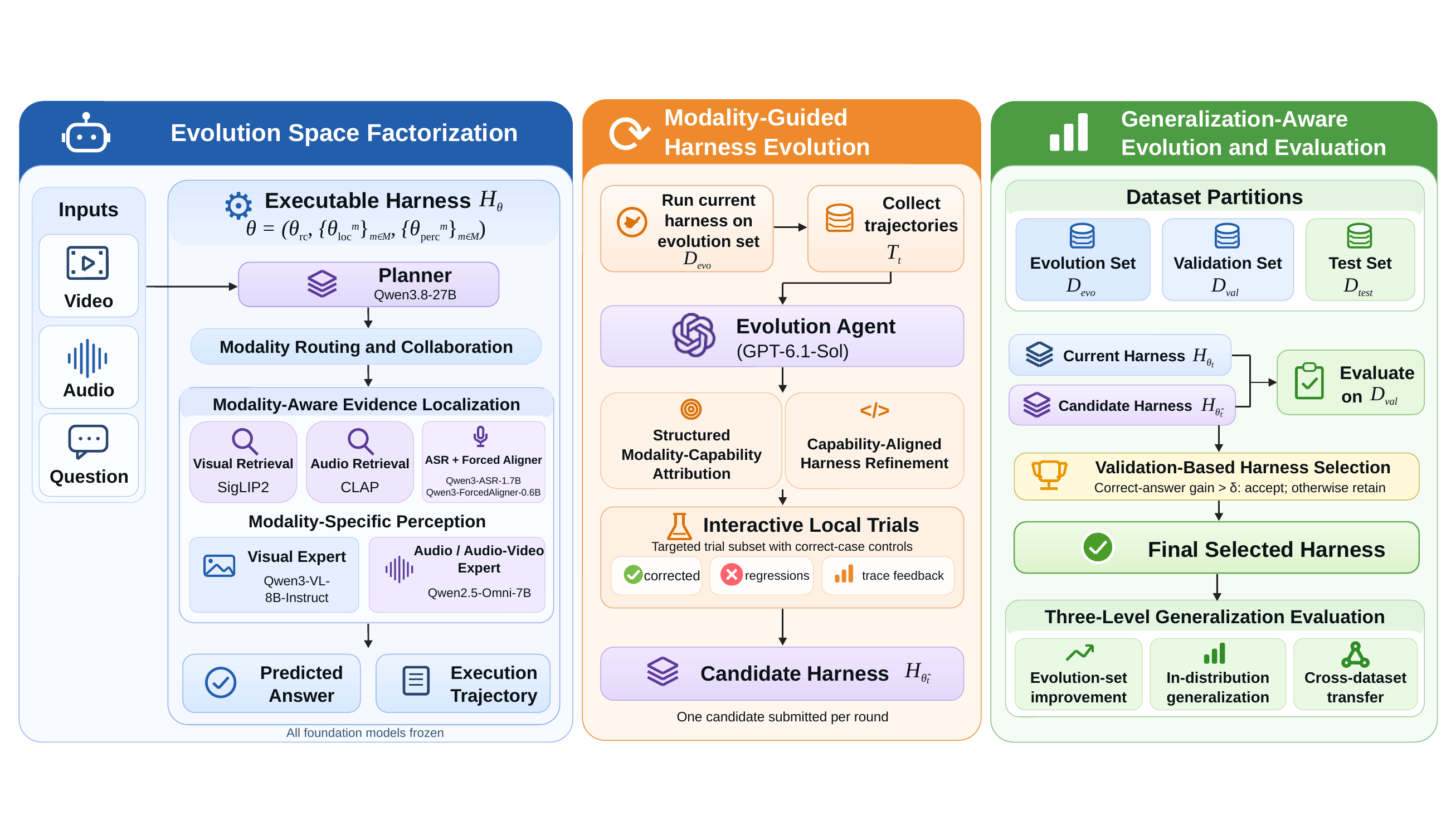}
    \caption{\textbf{Overview of OmniHarness-RSI.}
    The Factorized-Harness Agent separates editable strategies into modality routing and collaboration, modality-aware evidence localization, and modality-specific perception, with foundation models frozen.
    Structured modality-capability attribution guides capability-aligned harness refinement through interactive local trials.
    Validation-based harness selection accepts or rejects the single candidate submitted each round.
    The final selected harness undergoes three-level generalization evaluation: evolution-set improvement, in-distribution generalization, and cross-dataset transfer.}
    \label{fig:framework}
\end{figure*}

\subsection{Overview and Problem Formulation}
\label{sec:overview}

We propose \textbf{OmniHarness-RSI}, a modality-aware recursive self-improvement framework that systematically evolves executable agent harnesses for omnimodal understanding. Our framework improves agent capabilities by optimizing executable perception and reasoning workflows while keeping the parameters of all underlying foundation models frozen. As illustrated in Fig.~\ref{fig:framework}, OmniHarness-RSI consists of three key components: (i) a Factorized-Harness Agent that organizes editable harness strategies into three complementary capability dimensions, (ii) modality-guided harness evolution that combines structured modality-capability attribution with capability-aligned harness refinement, and (iii) a generalization-aware selection and evaluation protocol that assesses transferable improvements beyond evolution samples.

Formally, given an omnimodal understanding dataset
$\mathcal{D}=\{(v_i,a_i,q_i,y_i)\}_{i=1}^{N}$,
where $v_i$, $a_i$, $q_i$, and $y_i$ denote the video, audio, question, and ground-truth answer, respectively, we consider an agent harness $\mathcal{H}_{\theta}$ that orchestrates a collection of pretrained perception experts $\mathcal{E}=\{E_1,\ldots,E_K\}$. Here, $\theta$ represents the executable harness configuration, including tool invocation, perception strategies, evidence retrieval, and expert collaboration. The agent produces its prediction through
\begin{equation}
    \hat{y}_i =
    \mathcal{H}_{\theta}(v_i,a_i,q_i;\mathcal{E}).
    \label{eq:agent}
\end{equation}
Our objective is to progressively optimize $\theta$ to improve task performance and generalization across diverse omnimodal understanding scenarios.

\subsection{Modality-Aware Evolution Space Factorization}
\label{sec:omni_agent}

Omnimodal understanding involves intertwined capabilities for visual, auditory, and cross-modal reasoning. The harness strategies supporting these capabilities interact, creating an entangled evolution space that makes it difficult to attribute failures to specific strategies and target revisions accordingly. To address this challenge, we introduce \textbf{modality-aware evolution space factorization}, which organizes editable harness strategies into three complementary capability dimensions:

\noindent\textbf{Modality Routing and Collaboration.}
This dimension governs which modalities and experts to use for a question, in what order, and how to combine their observations and resolve conflicting evidence.

\noindent\textbf{Modality-Aware Evidence Localization.}
This dimension governs how to search different modalities for question-relevant evidence and locate the corresponding video frames or audio segments.

\noindent\textbf{Modality-Specific Perception.}
This dimension governs how visual, audio, and audio-video experts interpret the selected evidence to extract the details needed to answer the question.

Based on these three capability dimensions, we represent the executable harness configuration as
\begin{equation}
    \theta = \left(
        \theta_{\mathrm{rc}},
        \{\theta_{\mathrm{loc}}^{m}\}_{m\in\mathcal{M}},
        \{\theta_{\mathrm{perc}}^{m}\}_{m\in\mathcal{M}}
    \right),
    \label{eq:harness_decomposition}
\end{equation}
where $\theta_{\mathrm{rc}}$, $\theta_{\mathrm{loc}}^{m}$, and $\theta_{\mathrm{perc}}^{m}$ denote routing and collaboration, evidence localization, and perception strategies, respectively, and $\mathcal{M}$ indexes supported modalities or modality combinations. These are editable harness strategies; all foundation model parameters remain frozen.

Based on this factorization, we design a Factorized-Harness Agent that maps each capability dimension to concrete models, tools, and editable harness components. We use Qwen3.8-27B as the planner to decompose the question into evidence needs, select localization tools and perception experts, and integrate their observations into an answer. Given question $q$ and accumulated context $h_{s-1}$, it selects an expert or tool $e_s$ and an operation $u_s$:
\begin{equation}
    (e_s,u_s) = \pi_{\theta_{\mathrm{rc}}}(q,h_{s-1}),
    \label{eq:expert_interaction}
\end{equation}
where $\pi_{\theta_{\mathrm{rc}}}$ is the harness-governed routing and collaboration policy. Its editable components include planner prompts, tool availability and invocation rules, and logic for passing expert observations to the planner. Revising these components changes modality selection, call ordering, and policies for accepting, cross-checking, or overriding conflicting observations.

For evidence localization, we use SigLIP2~\cite{tschannen2025siglip2} for visual retrieval and CLAP~\cite{wu2022clap} for audio retrieval, together with Qwen3-ASR-1.7B and Qwen3-ForcedAligner-0.6B~\cite{shi2026qwen3asr} for time-aligned speech transcripts. The corresponding harness strategies $\theta_{\mathrm{loc}}^{m}$ include query-construction prompts, retrieval parameters such as candidate counts, and logic for combining, filtering, reranking, and deduplicating candidates. Transcript-based localization can be refined through keyword queries, transcript search strategies, and rules that map speech anchors to relevant time intervals. These components determine where to seek evidence through the available tools.

For perception, Qwen3-VL-8B-Instruct~\cite{bai2025qwen3vl} serves as the visual expert, while Qwen2.5-Omni-7B~\cite{xu2025qwen25omni} handles audio and joint audio-video understanding. Their harness strategies $\theta_{\mathrm{perc}}^{m}$ include task prompts, input organization, interface-supported frame counts and sampling density within selected intervals, and output formats and processing. For example, revisions can direct the visual expert to identify a target attribute or the audio expert to distinguish a target sound from background music. Truncation, concatenation, and simple parsing of returned text further shape the observations available to the planner.

The three dimensions organize strategies within a shared executable harness, allowing coordinated revisions across dimensions. We use GPT-6.1-Sol as the outer evolution agent to refine this harness through the modality-guided evolution process described next.

\subsection{Modality-Guided Harness Evolution}
\label{sec:interactive_evolution}

While modality-aware factorization organizes the evolution space into capability dimensions, this structure alone does not ensure effective optimization. Omnimodal understanding involves complex tool interactions and tightly coupled capabilities across modalities. Applying conventional evolution methods for video understanding~\cite{xu2026videoharnessrsi,luo2026videorsi} to these tasks can lead to two problems: \textit{ambiguous failure attribution}, where the modality capabilities limiting performance remain unclear, and \textit{multiple unfocused harness modifications}, which change several capabilities, making individual effects difficult to assess and subsequent refinement harder to target.

To address these challenges, we introduce \textbf{modality-guided harness evolution}, which comprises the following two components.

\noindent\textbf{Structured Modality-Capability Attribution.}
At evolution round $t$, we evaluate the current harness $\mathcal{H}_{\theta_t}$ on $\mathcal{D}_{\mathrm{evo}}$ to measure task accuracy and record per-example execution trajectories $\mathcal{T}_t=\{\tau_i\}$. Each trajectory contains tool calls, retrieved evidence, expert observations, reasoning, and the final prediction with its correctness label. For incorrect predictions, the GPT-6.1-Sol evolution agent analyzes the corresponding trajectories using the factorized space in Sec.~\ref{sec:omni_agent}, identifying likely deficiencies in routing and collaboration, evidence localization, or perception.

For routing and collaboration, attribution identifies the execution steps involving inappropriate modality selection, expert invocation order, evidence exchange, or conflict resolution. Within localization and perception, the agent performs \textit{modality--capability attribution} along two axes: the affected modality (e.g., visual or audio) and the specific editable harness strategy implicated in the failure. Here, capability refers to a fine-grained strategy, such as retrieval query construction, frame sampling within selected intervals, or expert prompting. For example, a diagnosis may associate the visual modality with retrieval query construction or frame sampling, or the audio modality with expert prompting. A trajectory may implicate multiple interacting strategies across these dimensions.

This process augments binary task outcomes with structured attribution results grounded in execution evidence and linked to concrete harness revision targets. These results guide subsequent editing and are assessed through interactive local trials.

\noindent\textbf{Capability-Aligned Harness Refinement.}
In practice, we find that attempting to address all identified failure patterns in a single harness revision yields limited improvement. We therefore prioritize focused revisions targeting specific failure patterns. For a selected pattern, the GPT-6.1-Sol evolution agent selects questions from $\mathcal{D}_{\mathrm{evo}}$ to form a trial subset, including cases exhibiting the pattern and related cases that the current harness answers correctly. This subset supports targeted assessment of the proposed changes while providing controls for detecting regressions.

Guided by the attribution results, the agent modifies the harness strategies associated with the selected pattern and evaluates candidate revisions on this subset. Comparing the current and revised harnesses on the same questions reveals which errors are corrected and whether previously correct predictions deteriorate. The agent also examines execution traces to assess how the revised strategies affect evidence acquisition and interpretation. It uses this feedback to refine the implementation and explore alternative revisions through repeated local trials. A revision may involve multiple components when they jointly contribute to the same failure pattern, and the trial subset can be adjusted as the analysis reveals additional relevant cases.

This process couples targeted strategy modification with execution-based feedback on relevant examples. At the end of each evolution round, GPT-6.1-Sol selects the candidate harness it judges to perform best based on the local trial feedback and submits only this candidate for validation-based selection (Sec.~\ref{sec:generalization}).

\subsection{Generalization-Aware Evolution and Evaluation}
\label{sec:generalization}

Modality-guided harness evolution refines harness strategies through interactive local trials, but improvements on trial subsets may not generalize. We therefore complement capability-aligned harness refinement with a generalization-aware selection and evaluation protocol that separates validation-based harness selection from final generalization evaluation.

\noindent\textbf{Validation-Based Harness Selection.}
For each dataset used in evolution, we construct three disjoint partitions: $\mathcal{D}_{\mathrm{evo}}$, $\mathcal{D}_{\mathrm{val}}$, and $\mathcal{D}_{\mathrm{test}}$. To mitigate overfitting to the evolution set during iterative harness refinement, we base candidate acceptance on accuracy improvements over the current harness on the held-out validation set.

At evolution round $t$, the GPT-6.1-Sol evolution agent submits one candidate harness $\mathcal{H}_{\hat{\theta}_t}$ based on local trial feedback. We compare it with the current harness $\mathcal{H}_{\theta_t}$ retained from the preceding round on the same validation set. Let $C_{\mathrm{val}}(\theta)$ denote the number of validation questions correctly answered by $\mathcal{H}_{\theta}$. The harness configuration is updated as
\begin{equation}
    \theta_{t+1} =
    \begin{cases}
        \hat{\theta}_t, & C_{\mathrm{val}}(\hat{\theta}_t)-C_{\mathrm{val}}(\theta_t)>\delta, \\
        \theta_t, & \text{otherwise},
    \end{cases}
    \label{eq:harness_selection}
\end{equation}
where $\delta\geq 0$ is a predefined threshold measured in correctly answered questions. The candidate harness is accepted only if its net gain strictly exceeds $\delta$; otherwise, the current harness is retained. This criterion assesses candidate revisions beyond their trial subsets. Aggregate validation feedback supports harness selection; test samples remain excluded from harness refinement and selection.

\begin{table*}[t]
    \centering
    \caption{\textbf{Effectiveness of Modality-Guided RSI across three benchmarks.}
    We report accuracy (\%) on the evolution, validation, and test splits.
    Gain rows show changes in percentage points relative to each architecture's initial harness.
    Bold denotes the highest accuracy and largest gain in each column.}
    \label{tab:main_results}
    \small
    \setlength{\tabcolsep}{5pt}
    \renewcommand{\arraystretch}{1.15}
    \begin{tabular*}{\textwidth}{@{\extracolsep{\fill}}l*{9}{c}@{}}
        \toprule
        Method & \multicolumn{3}{c}{WorldSense} & \multicolumn{3}{c}{Daily-Omni} & \multicolumn{3}{c}{OmniVideoBench} \\
        \cmidrule(lr){2-4}\cmidrule(lr){5-7}\cmidrule(lr){8-10}
        & Evolve & Valid & Test & Evolve & Valid & Test & Evolve & Valid & Test \\
        \midrule
        Single-Expert Agent & 48.25 & 48.17 & 46.44 & 69.34 & 54.63 & 63.96 & 31.93 & 54.32 & 36.48 \\
        \multirow{2}{*}{\shortstack[l]{Single-Expert Agent\\+ Modality-Agnostic RSI}} & 43.18 & 48.84 & 44.88 & 66.36 & 57.41 & 59.82 & 29.97 & 58.02 & 38.97 \\
        & $-5.07$ & $+0.67$ & $-1.56$ & $-2.98$ & $+2.78$ & $-4.14$ & $-1.96$ & $+3.70$ & $+2.49$ \\
        \midrule
        Factorized-Harness Agent & 57.78 & 52.49 & 54.20 & 75.97 & 76.85 & 74.08 & 40.06 & 56.79 & 42.17 \\
        \multirow{2}{*}{\shortstack[l]{Factorized-Harness Agent\\+ Modality-Agnostic RSI}} & 58.83 & 55.48 & 54.55 & 72.08 & 78.70 & 75.61 & 39.50 & 58.02 & 45.02 \\
        & $+1.05$ & $+2.99$ & $+0.35$ & $-3.89$ & $+1.85$ & $+1.53$ & $-0.56$ & $+1.23$ & $+2.85$ \\
        \multirow{2}{*}{\shortstack[l]{Factorized-Harness Agent\\+ Modality-Guided RSI}} & \textbf{65.03} & \textbf{57.81} & \textbf{63.06} & \textbf{81.01} & \textbf{83.33} & \textbf{82.21} & \textbf{44.82} & \textbf{64.20} & \textbf{48.40} \\
        & $\boldsymbol{+7.25}$ & $\boldsymbol{+5.32}$ & $\boldsymbol{+8.86}$ & $\boldsymbol{+5.04}$ & $\boldsymbol{+6.48}$ & $\boldsymbol{+8.13}$ & $\boldsymbol{+4.76}$ & $\boldsymbol{+7.41}$ & $\boldsymbol{+6.23}$ \\
        \bottomrule
    \end{tabular*}
\end{table*}

\noindent\textbf{Three-Level Generalization Evaluation.}
We assess the final selected harness at three complementary levels.

\textit{(i) Evolution-set improvement.}
Performance gains on $\mathcal{D}_{\mathrm{evo}}$ characterize harness improvement on evolution samples used during iterative refinement.

\textit{(ii) In-distribution generalization.}
Evaluation on held-out $\mathcal{D}_{\mathrm{test}}$ measures whether evolved harness strategies extend to unseen examples from the same dataset.

\textit{(iii) Cross-dataset transfer.}
Direct evaluation on datasets from different sources, without further adaptation or harness modification, assesses the transferability of evolved harness strategies across different data distributions.

At each level, we report gains over the initial harness under the corresponding benchmark metric. Together, these evaluations distinguish improvement on evolution samples from generalization to unseen examples and transfer across datasets.

\section{Experiments}
\label{sec:experiments}

\subsection{Experimental Setup}
\label{sec:experimental_setup}

\noindent\textbf{Datasets and metrics.}
We evaluate on three multiple-choice audio-video question-answering benchmarks: WorldSense~\cite{hong2026worldsense}, Daily-Omni~\cite{zhou2025dailyomni}, and OmniVideoBench~\cite{li2026omnivideobench}. They cover complementary aspects of omnimodal understanding, including visual, auditory, temporal, and cross-modal reasoning in WorldSense; everyday event understanding and audio-visual alignment in Daily-Omni; and 13 perception and reasoning categories in OmniVideoBench. We randomly partition each dataset into evolution, validation, and test sets in an approximate ratio of $40{:}10{:}50$. Performance is measured by answer accuracy (\%), with gains reported in percentage points.

\noindent\textbf{Agent configurations.}
We consider two agent architectures sharing the same Qwen3.8-27B planner. The \textit{Single-Expert Agent} employs Qwen2.5-Omni-7B as its sole perception expert, operating within an unfactorized harness evolution space where heterogeneous modality capabilities remain entangled. In contrast, our \textit{Factorized-Harness Agent} integrates multiple modality-specific experts and retrieval tools, and organizes the evolution space into three modality-related capability dimensions, as described in Sec.~\ref{sec:omni_agent}. For both architectures, the initial harness exposes predefined tool interfaces and provides basic tool-use instructions that guide the planner in selecting and invoking tools to answer questions.

\noindent\textbf{Evolution and evaluation protocol.}
We use GPT-6.1-Sol as the evolution agent and allow up to 10 evolution rounds per experiment. Candidate acceptance follows Eq.~\ref{eq:harness_selection}; evolution terminates early if the submitted candidate harness is rejected in three consecutive rounds. All model parameters remain frozen, and the planner is limited to 10 interaction steps per question at inference time.

We compare our \textit{Modality-Guided RSI} (Sec.~\ref{sec:interactive_evolution}) with
\textit{Modality-Agnostic RSI}, adapted from prior video harness evolution method~\cite{xu2026videoharnessrsi}. The latter follows a general-purpose harness evolution paradigm, optimizing executable strategies based on execution trajectories and evaluation feedback. In contrast, our approach explicitly leverages the factorized modality-capability space to identify specific deficiencies and guide subsequent harness refinement through execution feedback.

\subsection{Main Results}
\label{sec:main_results}

\noindent\textbf{Effectiveness of harness evolution.}
Table~\ref{tab:main_results} compares different agent architectures and harness evolution strategies across three benchmarks. For the Single-Expert Agent, Modality-Agnostic RSI struggles to achieve consistent improvements, reducing test accuracy by 1.56 and 4.14 percentage points on WorldSense and Daily-Omni, respectively, with only a 2.49-point gain on OmniVideoBench. This suggests that directly evolving an entangled harness has limited effectiveness in optimizing heterogeneous omnimodal capabilities.

Introducing the Factorized-Harness Agent improves the effectiveness of Modality-Agnostic RSI, yielding positive test gains of 0.35, 1.53, and 2.85 points across the three benchmarks. However, these improvements remain modest, and evolution-set accuracy still decreases by 3.89 and 0.56 points on Daily-Omni and OmniVideoBench. These results indicate that factorizing the evolution space provides a better foundation for harness optimization, but factorization alone is insufficient to fully exploit the structured capability space.

In contrast, our Modality-Guided RSI achieves substantial and consistent improvements, increasing test accuracy by 8.86, 8.13, and 6.23 points over the initial Factorized-Harness Agent, and outperforming Modality-Agnostic RSI by 8.51, 6.60, and 3.38 points, respectively. Moreover, improvements are observed across all evolution, validation, and test splits. These results demonstrate the effectiveness of combining modality-aware evolution space factorization with modality-guided harness refinement, enabling more focused optimization of modality-specific perception and cross-modal collaboration.

Notably, even before evolution, the Factorized-Harness Agent outperforms the Single-Expert Agent by 7.76, 10.12, and 5.69 points on the three test sets, highlighting the benefits of the factorized architecture and additional specialized perception and retrieval tools.

\begin{table}[t]
    \centering
    \caption{\textbf{Harness generalization across datasets.}
    Accuracy (\%) of the Factorized-Harness Agent with Modality-Guided RSI on each target dataset or transfer from WorldSense.}
    \label{tab:harness_generalization}
    \footnotesize
    \setlength{\tabcolsep}{3pt}
    \renewcommand{\arraystretch}{1.2}
    \begin{tabular*}{\columnwidth}{@{\extracolsep{\fill}}lcccc@{}}
        \toprule
        Method & \multicolumn{2}{c}{Daily-Omni} & \multicolumn{2}{c}{OmniVideoBench} \\
        \cmidrule(lr){2-3}\cmidrule(lr){4-5}
        & Valid & Test & Valid & Test \\
        \midrule
        Factorized-Harness Agent & 76.85 & 74.08 & 56.79 & 42.17 \\
        + Target-evolved harness & \textbf{83.33} & \textbf{82.21} & \textbf{64.20} & 48.40 \\
        + WorldSense-evolved harness & 80.56 & 81.75 & \textbf{64.20} & \textbf{54.27} \\
        \bottomrule
    \end{tabular*}
\end{table}

\noindent\textbf{Cross-dataset generalization.}
As shown in Table~\ref{tab:main_results}, Modality-Guided RSI achieves consistent improvements on both evolution and held-out test sets, demonstrating effective harness optimization and in-distribution generalization. We further investigate cross-dataset transfer in Table~\ref{tab:harness_generalization} by directly applying the WorldSense-evolved harness to Daily-Omni and OmniVideoBench without additional evolution.

Compared with the initial Factorized-Harness Agent, the transferred harness improves test accuracy by 7.67 and 12.10 percentage points on Daily-Omni and OmniVideoBench, respectively. Notably, its performance is comparable to that of harnesses evolved directly on the target datasets, achieving 81.75\% versus 82.21\% on Daily-Omni and even outperforming target-specific evolution on OmniVideoBench (54.27\% versus 48.40\%). These results demonstrate that Modality-Guided RSI learns transferable harness strategies that generalize across different omnimodal understanding benchmarks, rather than merely optimizing dataset-specific execution patterns.

\begin{table}[t]
    \centering
    \caption{Accuracy (\%) on Daily-Omni and OmniVideoBench. OmniAgent (matched) uses the same planner and perception models as our system; other baseline scores are reported in the cited papers. Our WorldSense-evolved harness is frozen on both target datasets.}
    \label{tab:comparison}
    \small
    \setlength{\tabcolsep}{3pt}
    \renewcommand{\arraystretch}{1.12}
    \begin{tabular*}{\columnwidth}{@{\extracolsep{\fill}}lrr@{}}
        \toprule
        Method & Daily-Omni & OmniVideoBench \\
        \midrule
        Qwen3-Omni-30B~\cite{tao2025omniagent} & 72.08 & 38.40 \\
        SyncRA~\cite{xu2026syncra} & & \\
        \quad Qwen3-Omni-30B-A3B & 79.09 & 47.07 \\
        OmniAgent (o3, reported)~\cite{tao2025omniagent} & 82.71 & 59.10 \\
        \midrule
        OmniAgent (matched)~\cite{tao2025omniagent} & 71.68 & 41.90 \\
        \textbf{OmniHarness-RSI (ours)} & 82.12 & 54.00 \\
        \bottomrule
    \end{tabular*}
\end{table}

\noindent\textbf{Comparison with existing methods.}

Table~\ref{tab:comparison} compares omnimodal models and agents on the full Daily-Omni and OmniVideoBench benchmarks. Our WorldSense-evolved harness achieves 82.12\% and 54.00\% accuracy without target-specific evolution. OmniAgent~\cite{tao2025omniagent} adopts a similar multi-expert architecture but uses stronger closed-source models. For a fair comparison, we reproduce it using the same planner and perception models as ours. OmniHarness-RSI outperforms this reproduction by 10.44 and 12.10 percentage points, respectively, supporting the benefits of our factorized harness design and Modality-Guided RSI under matched model configurations.

Compared with Qwen3-Omni-30B, OmniHarness-RSI improves accuracy by 10.04 and 15.60 points on Daily-Omni and OmniVideoBench, respectively. It also surpasses SyncRA with Qwen3-Omni-30B-A3B~\cite{xu2026syncra} by 3.03 and 6.93 points. These gains demonstrate that an evolved harness can effectively coordinate smaller perception experts to outperform a larger unified omnimodal model.

\subsection{Further Analysis}
\label{sec:further_analysis}

\begin{table}[t]
    \centering
    \caption{\textbf{Inference resource usage before and after harness evolution.}
    All values are per-question averages for the Factorized-Harness Agent. Initial and Evolved denote its harnesses before and after Modality-Guided RSI, respectively. Tool calls include perception experts, ASR, and audio/visual localization tools. Bold indicates lower usage within each dataset.}
    \label{tab:efficiency}
    \footnotesize
    \setlength{\tabcolsep}{3pt}
    \renewcommand{\arraystretch}{1.15}
    \begin{tabular*}{\columnwidth}{@{\extracolsep{\fill}}llrrr@{}}
        \toprule
        Dataset & Harness & \shortstack{Planner\\calls} & \shortstack{Tool\\calls} & \shortstack{Viewed\\frames} \\
        \midrule
        \multirow{2}{*}{WorldSense}
        & Initial & 5.32 & \textbf{6.71} & 177.93 \\
        & Evolved & \textbf{4.40} & 7.22 & \textbf{171.02} \\
        \midrule
        \multirow{2}{*}{Daily-Omni}
        & Initial & 4.94 & 6.52 & 129.39 \\
        & Evolved & \textbf{2.92} & \textbf{4.49} & \textbf{110.70} \\
        \midrule
        \multirow{2}{*}{OmniVideoBench}
        & Initial & 6.83 & \textbf{8.90} & 217.64 \\
        & Evolved & \textbf{6.45} & 9.75 & \textbf{207.35} \\
        \bottomrule
    \end{tabular*}
\end{table}

\noindent\textbf{Inference Efficiency.}
To examine whether the performance gains of Modality-Guided RSI require additional inference resources, Table~\ref{tab:efficiency} compares planner calls, tool invocations, and viewed frames before and after evolution. Despite achieving higher accuracy across all three benchmarks, the evolved harness reduces planner calls by 17.3\%, 40.9\%, and 5.6\% on WorldSense, Daily-Omni, and OmniVideoBench, respectively, while reducing viewed frames by 3.9\%, 14.4\%, and 4.7\%. On Daily-Omni, tool calls also decrease by 31.1\%, achieving simultaneous improvements across all three resource metrics. Although tool calls increase slightly on the other two benchmarks, the consistent reductions in planning interactions and visual inputs indicate that accuracy gains do not depend on uniformly increased resource consumption. Instead, the evolved harness demonstrates more effective allocation of planning and perception resources, enabling more efficient evidence acquisition and utilization.

\begin{figure}[t]
    \centering
    \includegraphics[width=\columnwidth]{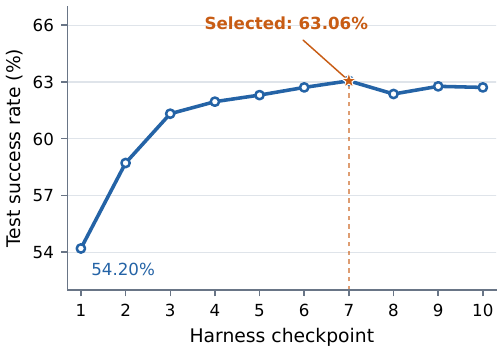}
    \caption{\textbf{WorldSense test performance during Modality-Guided RSI.}
    Checkpoints are shown in chronological order.
    The orange star marks the final adopted harness (checkpoint 7), which achieves 63.06\%, a gain of 8.86 percentage points over the first checkpoint (54.20\%).}
    \label{fig:worldsense_evolution}
\end{figure}

\noindent\textbf{Evolution Dynamics.}
Figure~\ref{fig:worldsense_evolution} illustrates the performance trajectory of Modality-Guided RSI across ten successive harness checkpoints on WorldSense. The test success rate rises steadily during the early stages of evolution, increasing from 54.20\% to 63.06\% at checkpoint 7, an 8.86-percentage-point gain. This trend suggests that Structured Modality-Capability Attribution identifies relevant failure types and guides Capability-Aligned Harness Refinement to progressively address them through targeted harness revisions. Subsequent checkpoints fluctuate around 62.36\%--62.77\%, indicating that further refinement does not necessarily yield additional improvements. The final harness is selected using the validation-based criterion in Sec.~\ref{sec:generalization}.

\begin{figure}[t]
    \centering
    \includegraphics[width=\columnwidth]{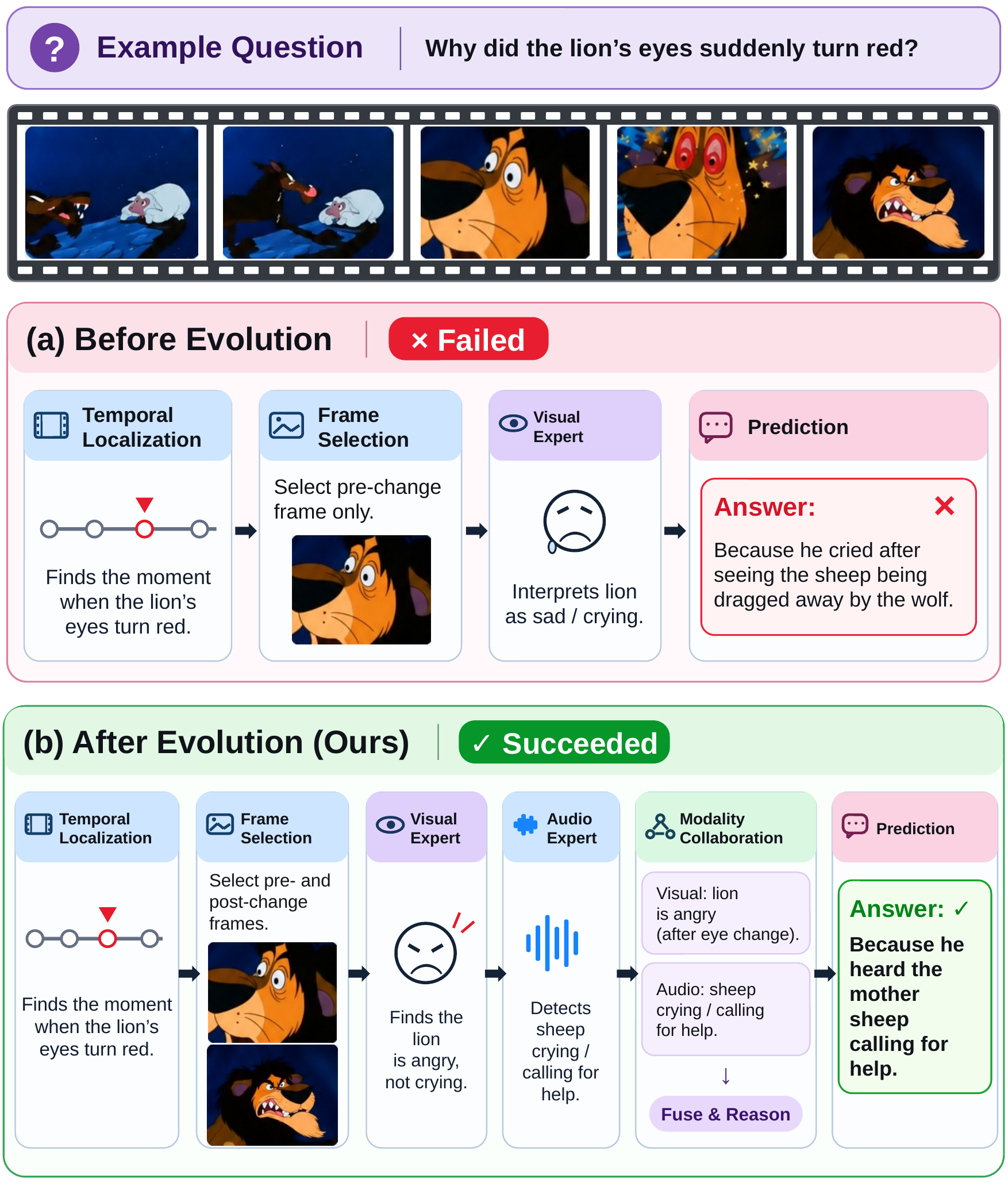}
    \caption{\textbf{Qualitative comparison before and after Modality-Guided RSI.}
    The question asks why the lion's eyes suddenly turn red.
    Before evolution, the harness selects only a pre-change frame, and the visual expert interprets the lion as sad or crying, leading to an incorrect explanation.
    After evolution, the harness selects frames both before and after the change, identifies the lion's anger, and incorporates audio evidence of the mother sheep calling for help.
    Combining the visual and auditory evidence yields the correct answer in this example.}
    \label{fig:qualitative_analysis}
\end{figure}

\noindent\textbf{Qualitative Analysis of Evolved Harnesses.}
Figure~\ref{fig:qualitative_analysis} illustrates how Modality-Guided RSI improves visual sampling around a localized event and collaboration between visual and audio experts. For the question ``Why did the lion's eyes suddenly turn red?'', the initial harness correctly localizes the event but selects only a frame preceding the change. This leaves the visual expert without evidence of the post-change state, leading it to interpret the lion as sad or crying and produce an incorrect causal explanation. The failure illustrates that accurate temporal localization alone is insufficient when subsequent sampling omits the evidence needed to interpret the event.

After evolution, the harness uses the localized event as a temporal anchor and samples frames both before and after the change. This strategy provides the visual expert with temporal context for interpreting the change in expression and recognizing the lion's anger. The evolved harness also coordinates complementary visual and auditory observations: the visual expert identifies the lion's emotional response, while the audio expert detects the sheep crying or calling for help. By integrating these observations, the planner links the visible reaction to its auditory cause and correctly attributes the eye-color change to hearing the mother sheep calling for help. This case illustrates how harness evolution improves the use of localization results for visual sampling and the integration of expert observations for cross-modal causal reasoning, while keeping the perception models frozen.

\section{Conclusion}
\label{sec:conclusion}

We presented OmniHarness-RSI, a modality-aware framework for improving omnimodal understanding through executable harness evolution while keeping foundation model parameters frozen. Our Factorized-Harness Agent organizes the evolution space into three complementary dimensions: Modality Routing and Collaboration, Modality-Aware Evidence Localization, and Modality-Specific Perception. This factorization provides explicit targets for optimizing expert coordination, evidence acquisition, and interpretation. Building on this structure, Modality-Guided RSI combines Structured Modality-Capability Attribution with Capability-Aligned Harness Refinement to connect failure diagnosis with targeted code revisions and execution feedback. A generalization-aware protocol separates evolution, validation-based harness selection, and held-out evaluation. Experiments on WorldSense, Daily-Omni, and OmniVideoBench show test accuracy gains of 8.86, 8.13, and 6.23 percentage points over the initial Factorized-Harness Agent, consistently outperforming Modality-Agnostic RSI. The WorldSense-evolved harness also transfers to both target datasets without further adaptation. These findings support structured harness evolution for improving how frozen experts acquire, interpret, and integrate multimodal evidence. Future work will examine evolution efficiency and robustness across backbones, tasks, and data distributions.


{
    \small
    \bibliographystyle{ieeenat_fullname}
    \bibliography{main}

@article{xu2026videoharnessrsi,
  title={VideoHarness-RSI: Recursive Harness Self-Improvement for Long-Video Understanding with Frozen Vision-Language Models},
  author={Xu, Guoyang and Chen, Hao},
  journal={arXiv preprint arXiv:2608.24302},
  year={2026}
}

@article{luo2026videorsi,
  title={Video-RSI: Recursive Self-Improvement of Video Understanding Agents via Harness Evolution},
  author={Luo, Bingjun and Guo, Jialin and Li, Siqi},
  journal={arXiv preprint arXiv:2609.37950},
  year={2026}
}

@article{xu2026videoevoharness,
  title={VideoEvoHarness: Self Evolving Agent Harness from Unlabeled Videos for Video Understanding},
  author={Xu, Weixin and Liu, Yifan and Wang, Yi and Rong, Mingfei and Yang, Zhenyu and Zhong, Zhisheng and Qian, Shengsheng and Xu, Changsheng},
  journal={arXiv preprint arXiv:2610.15389},
  year={2026}
}

@article{lin2026videoharnesssrsi,
  title={VideoHarness-SRSI: Structured Recursive Self-Improvement for Video Understanding Harness},
  author={Lin, Zihao and Dong, Haoyu and Xu, Haobo and Chang, Chi-Chih and Ning, Xuying and Xia, Yinglong and Yang, Shuang and Zhao, Zhuokai and Song, Xinying and Liu, Dongfang and Huang, Lifu and Song, Yang and Wang, Qifan},
  journal={arXiv preprint arXiv:2610.19424},
  year={2026}
}

@article{tao2025omniagent,
  title={OmniAgent: Audio-Guided Active Perception Agent for Omnimodal Audio-Video Understanding},
  author={Tao, Keda and Du, Wenjie and Yu, Bohan and Wang, Weiqiang and Liu, Jian and Wang, Huan},
  journal={arXiv preprint arXiv:2512.23646},
  year={2025}
}

@article{xu2025qwen3omni,
  title={Qwen3-Omni Technical Report},
  author={Xu, Jin and others},
  journal={arXiv preprint arXiv:2509.17765},
  year={2025}
}

@article{cheng2024videollama2,
  title={VideoLLaMA 2: Advancing Spatial-Temporal Modeling and Audio Understanding in Video-LLMs},
  author={Cheng, Zesen and Leng, Sicong and Zhang, Hang and Xin, Yifei and Li, Xin and Chen, Guanzheng and Zhu, Yongxin and others},
  journal={arXiv preprint arXiv:2406.07476},
  year={2024}
}

@inproceedings{sun2024videosalmonn,
  title={video-SALMONN: Speech-Enhanced Audio-Visual Large Language Models},
  author={Sun, Guangzhi and Yu, Wenyi and Tang, Changli and Chen, Xianzhao and Tan, Tian and Li, Wei and Lu, Lu and Ma, Zejun and Wang, Yuxuan and Zhang, Chao},
  booktitle={International Conference on Machine Learning},
  year={2024}
}

@inproceedings{sun2025videosalmonno1,
  title={video-SALMONN-o1: Reasoning-enhanced Audio-visual Large Language Model},
  author={Sun, Guangzhi and Yang, Yudong and Zhuang, Jimin and Tang, Changli and Li, Yixuan and Li, Wei and Ma, Zejun and Zhang, Chao},
  booktitle={International Conference on Machine Learning},
  year={2025}
}

@inproceedings{li2026omnivideobench,
  title={OmniVideoBench: Towards Audio-Visual Understanding Evaluation for Omni MLLMs},
  author={Li, Caorui and Chen, Yu and Ji, Yiyan and Xu, Jin and Cui, Zhenyu and Li, Shihao and others},
  booktitle={International Conference on Learning Representations},
  year={2026},
  eprint={2510.10689},
  archivePrefix={arXiv}
}

@inproceedings{wang2026avibench,
  title={AVI-Bench: Toward Human-like Audio-Visual Intelligence of Omni-MLLMs},
  author={Wang, Yaoting and Zhang, Ziyi and Tu, Wenming and Xu, Shaoxuan and Du, Wenjie and Liang, Cheng and others},
  booktitle={International Conference on Machine Learning},
  year={2026},
  eprint={2606.07643},
  archivePrefix={arXiv}
}

@article{chao2026jointavbench,
  title={JointAVBench: A Benchmark for Joint Audio-Visual Reasoning Evaluation},
  author={Chao, Jianghan and Gao, Jianzhang and Tan, Wenhui and Sun, Yuchong and Song, Ruihua and Ru, Liyun},
  journal={arXiv preprint arXiv:2512.12772},
  year={2026}
}

@article{chen2026omnireasoner,
  title={{OmniReasoner: Thinking with Long Audio-Video via Native Tool Use}},
  author={Chen, Yu and Li, Caorui and Xiong, Ziyu and Wang, Yidong and Gao, Mingqi and Liu, Shuman and Liu, Biao and Yang, Chunfeng and Zeng, Anxiang and Zhang, Haibo and Chen, Chaofan},
  journal={arXiv preprint arXiv:2607.19339},
  year={2026}
}

@article{jia2026omnijigsaw,
  title={{OmniJigsaw: Enhancing Omni-Modal Reasoning via Modality-Orchestrated Reordering}},
  author={Jia, Yiduo and Zhu, Muzhi and Zhong, Hao and Liu, Mingyu and Xi, Yuling and Chen, Hao and Qin, Bin and Yang, Yongjie and Luo, Zhenbo and Shen, Chunhua},
  journal={arXiv preprint arXiv:2604.08209},
  year={2026}
}

@article{chen2026omnivideor1,
  title={{OmniVideo-R1: Reinforcing Audio-visual Reasoning with Query Intention and Modality Attention}},
  author={Chen, Zhangquan and Tao, Jiale and Li, Ruihuang and Hu, Yihao and Chen, Ruitao and Yang, Zhantao and Yu, Xinlei and Jing, Haodong and Zhang, Manyuan and Shao, Shuai and Wang, Biao and Lu, Qinglin and Huang, Ruqi},
  journal={arXiv preprint arXiv:2602.05847},
  year={2026}
}

@article{liu2026omniboost,
  title={{Boosting Omni-Modal Language Models: Staged Post-Training with Visually Debiased Evaluation}},
  author={Liu, Che and Ma, Lichao and Zhang, Xiangyu Tony and Zhang, Yuxin and Zhang, Haoyang and Yang, Xuerui and Tian, Fei},
  journal={arXiv preprint arXiv:2605.12034},
  year={2026}
}

@article{xu2026syncra,
  title={{SyncRA: Learning Temporal Correspondence in Omni-Modal Models}},
  author={Xu, Zelong and Li, Yan and Hu, Wenhe and Hu, Xiyang},
  journal={arXiv preprint arXiv:2609.34363},
  year={2026}
}

@article{wu2026sleuthvlm,
  title={{SleuthVLM: Do We Really Need Audio-in-Video for Omni Video Understanding?}},
  author={Wu, Peiran and Gou, Fangfei and Mu, Zhaoxi and Zhang, Changhao and Liu, Jian and Wang, Weiqiang and Wang, Yuntao and Shi, Yuanchun and Shen, Junxiao},
  journal={arXiv preprint arXiv:2611.06083},
  year={2026}
}

@article{xu2025qwen25omni,
  title={{Qwen2.5-Omni Technical Report}},
  author={Xu, Jin and Guo, Zhifang and He, Jinzheng and Hu, Hangrui and He, Ting and Bai, Shuai and Chen, Keqin and Wang, Jialin and Fan, Yang and Dang, Kai and Zhang, Bin and Wang, Xiong and Chu, Yunfei and Lin, Junyang},
  journal={arXiv preprint arXiv:2503.20215},
  year={2025}
}

@inproceedings{hong2026worldsense,
  title={{WorldSense}: Evaluating Real-world Omnimodal Understanding for Multimodal {LLMs}},
  author={Hong, Jack and Yan, Shilin and Cai, Jiayin and Jiang, Xiaolong and Hu, Yao and Xie, Weidi},
  booktitle={International Conference on Learning Representations},
  year={2026}
}

@article{zhou2025dailyomni,
  title={{Daily-Omni}: Towards Audio-Visual Reasoning with Temporal Alignment across Modalities},
  author={Zhou, Ziwei and Wang, Rui and Wu, Zuxuan and Jiang, Yu-Gang},
  journal={arXiv preprint arXiv:2505.17862},
  year={2025}
}

@article{tschannen2025siglip2,
  title={{SigLIP 2}: Multilingual Vision-Language Encoders with Improved Semantic Understanding, Localization, and Dense Features},
  author={Tschannen, Michael and Gritsenko, Alexey and Wang, Xiao and Naeem, Muhammad Ferjad and Alabdulmohsin, Ibrahim and Parthasarathy, Nikhil and Evans, Talfan and Beyer, Lucas and Xia, Ye and Mustafa, Basil and H{\'e}naff, Olivier and Harmsen, Jeremiah and Steiner, Andreas and Zhai, Xiaohua},
  journal={arXiv preprint arXiv:2502.14786},
  year={2025}
}

@article{wu2022clap,
  title={Large-scale Contrastive Language-Audio Pretraining with Feature Fusion and Keyword-to-Caption Augmentation},
  author={Wu, Yusong and Chen, Ke and Zhang, Tianyu and Hui, Yuchen and Nezhurina, Marianna and Berg-Kirkpatrick, Taylor and Dubnov, Shlomo},
  journal={arXiv preprint arXiv:2211.06687},
  year={2022}
}

@article{shi2026qwen3asr,
  title={{Qwen3-ASR} Technical Report},
  author={Shi, Xian and Wang, Xiong and Guo, Zhifang and Wang, Yongqi and Zhang, Pei and Zhang, Xinyu and Guo, Zishan and Hao, Hongkun and Xi, Yu and Yang, Baosong and Xu, Jin and Zhou, Jingren and Lin, Junyang},
  journal={arXiv preprint arXiv:2601.21337},
  year={2026}
}

@article{bai2025qwen3vl,
  title={{Qwen3-VL} Technical Report},
  author={Bai, Shuai and Cai, Yuxuan and Chen, Ruizhe and Chen, Keqin and Chen, Xionghui and Cheng, Zesen and others},
  journal={arXiv preprint arXiv:2511.21631},
  year={2025}
}

@article{lee2026rhi,
  title={Recursive Harness Self-Improvement},
  author={Lee, Hyunin and Xu, Jinglue and Seely, Jeffrey and Lee, Donghyun and Zaharia, Matei and Tang, Yujin},
  journal={arXiv preprint arXiv:2607.15524},
  year={2026}
}

@article{lin2026ahe,
  title={Agentic Harness Engineering: Observability-Driven Automatic Evolution of Coding-Agent Harnesses},
  author={Lin, Jiahang and Liu, Shichun and Pan, Chengjun and Lin, Lizhi and Dou, Shihan and Xi, Zhiheng and Huang, Xuanjing and Yan, Hang and Han, Zhenhua and Gui, Tao and Jiang, Yu-Gang},
  journal={arXiv preprint arXiv:2604.25850},
  year={2026}
}

@article{wu2026modularrsi,
  title={{ModularRSI}: Modular and Generalizable Recursive Harness Self-Improvement},
  author={Wu, Siwei and Ren, Jincheng and Li, Yizhi and Li, Haau-Sing and Yang, Chengran and Ma, Xudong and Zhang, Yuxuan and Gu, Weicheng and Yang, Jian and Batista-Navarro, Riza and Zhang, Chuanyi and Liu, Xianglong and Zhou, Ming and Dai, Bryan and Lin, Chenghua},
  journal={arXiv preprint arXiv:2609.14857},
  year={2026}
}

@article{xia2026rrsi,
  title={{RRSI}: Regularized Recursive Self-Improvement of Agent Harnesses},
  author={Xia, Peng and Han, Rujun and Wang, Zifeng and Chen, Yanfei and Zhuang, Yufan and Lee, Yoonho and Huang, Chengsong and Yu, Han and CuiZhu, Zhongying and Ming, Yifei and Yao, Huaxiu and Gokturk, Burak and Pfister, Tomas and Lee, Chen-Yu},
  journal={arXiv preprint arXiv:2609.24972},
  year={2026}
}

@article{fu2024vita,
  title={{VITA}: Towards Open-Source Interactive Omni Multimodal {LLM}},
  author={Fu, Chaoyou and Lin, Haojia and Long, Zuwei and Shen, Yunhang and Dai, Yuhang and Zhao, Meng and Zhang, Yi-Fan and Dong, Shaoqi and Li, Yangze and Wang, Xiong and Cao, Haoyu and Yin, Di and Ma, Long and Zheng, Xiawu and Ji, Rongrong and Wu, Yunsheng and He, Ran and Shan, Caifeng and Sun, Xing},
  journal={arXiv preprint arXiv:2408.05211},
  year={2024}
}

@article{li2025baichuanomni15,
  title={{Baichuan-Omni-1.5} Technical Report},
  author={Li, Yadong and Liu, Jun and Zhang, Tao and Chen, Song and Li, Tianpeng and Li, Zehuan and others},
  journal={arXiv preprint arXiv:2501.15368},
  year={2025}
}

@article{xing2025echoinkr1,
  title={{EchoInk-R1}: Exploring Audio-Visual Reasoning in Multimodal {LLMs} via Reinforcement Learning},
  author={Xing, Zhenghao and Hu, Xiaowei and Fu, Chi-Wing and Wang, Wenhai and Dai, Jifeng and Heng, Pheng-Ann},
  journal={arXiv preprint arXiv:2505.04623},
  year={2025}
}

@inproceedings{li2025omnibench,
  title={{OmniBench}: Towards The Future of Universal Omni-Language Models},
  author={Li, Yizhi and Zhang, Ge and Ma, Yinghao and Yuan, Ruibin and Zhu, Kang and Guo, Hangyu and others},
  booktitle={Advances in Neural Information Processing Systems},
  volume={38},
  year={2025}
}

@article{tao2026lvomnibench,
  title={{LVOmniBench}: Pioneering Long Audio-Video Understanding Evaluation for Omnimodal {LLMs}},
  author={Tao, Keda and Zheng, Yuhua and Xu, Jia and Du, Wenjie and Shao, Kele and Wang, Hesong and Chen, Xueyi and Jin, Xin and Zhu, Junhan and Yu, Bohan and Wang, Weiqiang and Liu, Jian and Qin, Can and Zhang, Yulun and Yang, Ming-Hsuan and Wang, Huan},
  journal={arXiv preprint arXiv:2603.19217},
  year={2026}
}
}


\end{document}